\documentclass[11pt]{article}

\usepackage[T1]{fontenc}
\usepackage[utf8]{inputenc}
\usepackage{mathptmx}
\usepackage[margin=1in]{geometry}
\usepackage{microtype}
\usepackage{amsmath,amssymb}
\usepackage{graphicx}
\usepackage{booktabs}
\usepackage{tabularx}
\usepackage{array}
\usepackage{enumitem}
\usepackage{xcolor}
\usepackage{caption}
\usepackage{tikz}
\usetikzlibrary{arrows.meta,positioning,fit,backgrounds}
\usepackage[numbers,sort&compress]{natbib}
\usepackage[hidelinks]{hyperref}
\usepackage{url}

\setlist{itemsep=2pt,topsep=4pt}
\newcolumntype{L}{>{\raggedright\arraybackslash}X}
\newcommand{\code}[1]{\texttt{#1}}

\title{\textbf{Plant Memory: Technical Language Processing and a Verified Maintenance Ontology for Agent-Accessible Failure Knowledge from SAP PM Work Orders}}

\author{Akash Raj\\[2pt]
\small M.Tech, School of Artificial Intelligence and Data Science\\
\small Indian Institute of Technology Jodhpur\\
\small \href{mailto:m25ai1116@iitj.ac.in}{\texttt{m25ai1116@iitj.ac.in}}\\
\small ORCID: \href{https://orcid.org/0009-0005-5666-3967}{0009-0005-5666-3967}}

\date{October 2026}

\begin{document}
\maketitle

\begin{abstract}
Maintenance work orders are the largest written record of how industrial equipment fails, yet in most plants they are close to unusable for analysis. Short texts are abbreviated and misspelt, asset identity is buried in free text, work-type labels contradict the work described, and the fields that matter most for reliability (symptom, root cause, action taken) are usually empty. As a result, diagnostic knowledge stays with senior technicians and leaves when they do. We present \emph{Plant Memory}, a system that applies technical language processing (TLP) to SAP Plant Maintenance (PM) exports and builds an ISO~14224-aligned maintenance ontology in which every extracted statement is a reified fact carrying a confidence score, the evidence it was read from, and a verification state. Extraction is rule-first, with an optional large language model (LLM) acting as a second reader whose disagreements raise a fact's review priority rather than overwrite it. Technicians verify facts through a review queue, answer targeted knowledge-gap questions where failures repeat without a recorded cause, and approve or reject ontology drift proposals when new terms appear. The resulting graph is served to people through a web application and to AI agents through a Model Context Protocol (MCP) server, so both query the same verified knowledge and every answer cites the work orders behind it. On a synthetic 12-month export of 400 work orders from a parboiled-rice plant, rule-based extraction reaches 96.2\% component accuracy, 88.9\% failure-mode accuracy and 100\% root-cause precision at 62\% coverage; on 28 diagnostic questions, ranked root causes from the ontology are correct at top-1 in 82.1\% of cases against 53.6\% for the raw SAP root-cause column. We also report that an LLM used as an independent extractor was less accurate than the rules on every field, and describe the merge policy this led to. We discuss why these numbers are optimistic and what a real-plant evaluation requires.
\end{abstract}

\section{Introduction}

A typical computerised maintenance management system (CMMS) export records which parts were issued to which order and what they cost. It rarely records why the equipment failed or what actually fixed it. Where that information exists, it is spread across a 40-character short text (``Parb.- Dryer elevator brg chng''), an optional remarks field written in a mix of English and Hindi (``grease nahi tha, brg jam''), and occasionally a root-cause column filled in for a fraction of breakdowns. Reliability engineers who want to know which failure modes drive downtime, or which spares to stock, must read these records by hand.

Two developments make this worth revisiting. First, technical language processing has emerged as a discipline concerned with exactly this kind of text: short, domain-specific, abbreviation-heavy, and written for colleagues rather than for machines \citep{brundage2021tlp,dima2021adapting}. Second, AI assistants are starting to be placed in front of planners and engineers. An assistant can only answer from data it can read, and a raw Excel export is a poor substrate: it invites confident answers built on misread abbreviations, with no way for the reader to check them.

Plant Memory addresses both. Its contributions are:
\begin{enumerate}
  \item \textbf{A provenance-first maintenance ontology.} Classes and relations are aligned with ISO~14224 \citep{iso14224} and extended for rice-processing equipment. Every extracted statement is reified as a fact with confidence, extraction method, evidence snippet, source field and a verification state (proposed, verified, rejected).
  \item \textbf{A TLP pipeline tuned for SAP PM exports.} It merges material lines into events, expands plant abbreviations, resolves assets from functional locations or free-text aliases, and extracts component, symptom, failure mode, root cause and action under an explicit precedence order with calibrated confidences.
  \item \textbf{An LLM-as-second-reader merge policy}, derived from a measured comparison in which the LLM underperformed the rules as an independent extractor but added value by filling gaps and flagging disagreements.
  \item \textbf{A human verification loop} comprising a lowest-confidence-first review queue, automatically generated knowledge-gap questions for technicians, and ontology drift proposals that change the vocabulary only on approval. Verification decisions survive rebuilds.
  \item \textbf{One knowledge core behind three interfaces}: a web application, a command-line interface and an MCP server \citep{anthropic2024mcp}. The MCP server lets general-purpose AI agents query the verified graph and cite evidence, while its only write tool creates proposals.
  \item \textbf{An evaluation harness} with a synthetic SAP PM generator that plants known patterns and hidden ground truth, scoring both extraction and end-to-end diagnosis.
\end{enumerate}

\section{Background and related work}

\paragraph{Technical language processing.} \citet{brundage2021tlp} argue that general-purpose NLP pipelines fail on maintenance text because of its vocabulary, brevity and implicit context, and they propose TLP as a human-in-the-loop discipline that adapts NLP tools to technical domains. \citet{dima2021adapting} discuss the adaptations this requires. Earlier work on cleaning work-order data for reliability analysis \citep{hodkiewicz2016cleaning} and on hybrid human--machine tagging of maintenance logs \citep{sexton2017hybrid} established that useful structure can be recovered with modest, carefully designed tooling and that expert review is central. MaintIE \citep{bikaun2024maintie} provides a fine-grained annotation schema and benchmark for information extraction from maintenance short texts. Plant Memory follows the TLP view: rules encode plant-specific knowledge, and people remain responsible for what counts as true.

\paragraph{Maintenance ontologies.} ISO~14224 \citep{iso14224} standardises equipment taxonomies, failure modes and failure mechanisms for reliability data. Reference ontologies for industrial maintenance, such as ROMAIN \citep{karray2019romain} and the maintenance-activity ontology of \citet{woods2024ontology}, formalise these concepts with richer axiomatisation. Our schema is deliberately lighter (an OWL~2 \citep{owl2} vocabulary with reified facts) because its purpose is operational: to carry provenance and verification state for every statement extracted from a specific plant's records.

\paragraph{LLMs, knowledge graphs and tool protocols.} Retrieval-augmented generation \citep{lewis2020rag} grounds model output in retrieved text; graph-based variants \citep{edge2024graphrag} and broader proposals for unifying LLMs with knowledge graphs \citep{pan2024unifying} retrieve from structured knowledge instead. The Model Context Protocol \citep{anthropic2024mcp} standardises how assistants call external tools and read resources. Plant Memory uses MCP to expose reasoning tools over a verified graph rather than raw documents, so that an assistant's answer can be traced to specific work orders and verification states.

\section{Problem setting: what an SAP PM export looks like}
\label{sec:problem}

The input is an SAP PM export in Excel or CSV with one row per material issue. A month of a real export from a rice processing unit motivated the design and exhibits the following issues, which the synthetic data reproduces:
\begin{itemize}
  \item \textbf{Repeat failures are not linked.} One elevator accumulated three separate orders within a month (a bearing change, a sprocket change and ``general maintenance''), and nothing connects them as one recurring problem.
  \item \textbf{Free text is not machine-readable.} Descriptions mix abbreviations and typos (``Parb.'', ``brg'', ``Plmbr Blck'', ``modificatn'', ``ELevator''), and asset names (``Hodi Elevator'', ``Elev1\&2'') are embedded in text with no equipment identifier.
  \item \textbf{Work-type labels are unreliable.} Bearing and belt replacements are tagged \emph{Preventive} while comparable work is tagged \emph{Corrective}, which skews any analysis built on the label.
  \item \textbf{Cause and fix are missing.} Cost, quantity and material are present; symptom, root cause, downtime and action taken mostly are not.
  \item \textbf{Records are fragmented.} One order may be posted on several dates, materials are split across lines, and reservation rows (SAP transaction MB21) carry a cost centre but no order number.
\end{itemize}

\section{System design}

\subsection{Architecture}

\begin{figure}[t]
\centering
\begin{tikzpicture}[
  node distance=5mm and 7mm,
  box/.style={draw, rounded corners=2pt, align=center, font=\small, minimum height=10mm, inner sep=4pt, fill=white},
  store/.style={box, fill=black!6},
  arr/.style={-{Stealth[length=2mm]}, thick}]
  \node[box] (x) {SAP PM / CMMS\\export (xlsx, csv)};
  \node[box, right=of x] (u) {Ingest\\column mapping,\\line merge, assets};
  \node[box, right=of u] (n) {Normalise\\abbreviations,\\asset resolution};
  \node[box, right=of n] (e) {Extract\\rules first,\\optional LLM};
  \node[store, below=10mm of e] (g) {Knowledge graph\\Neo4j or SQLite\\reified facts};
  \node[box, left=14mm of g] (r) {Review queue,\\gap questions,\\drift proposals};
  \node[box, below=10mm of g, xshift=-38mm] (w) {Web app\\(people)};
  \node[box, below=10mm of g] (m) {MCP server\\(AI agents)};
  \node[box, below=10mm of g, xshift=38mm] (c) {CLI\\(automation)};
  \node[box, left=of r] (o) {RDF / Turtle\\export};
  \draw[arr] (x) -- (u); \draw[arr] (u) -- (n); \draw[arr] (n) -- (e); \draw[arr] (e) -- (g);
  \draw[arr] ([yshift=-2.5mm]g.west) -- node[below, font=\scriptsize]{proposals} ([yshift=-2.5mm]r.east);
  \draw[arr] ([yshift=2.5mm]r.east) -- node[above, font=\scriptsize]{decisions} ([yshift=2.5mm]g.west);
  \draw[arr] (g) -- (w); \draw[arr] (g) -- (m); \draw[arr] (g) -- (c);
  \draw[arr, dashed] (g.south west) to[bend left=18] (o.south);
\end{tikzpicture}
\caption{Plant Memory architecture. All three interfaces read and write the same graph through one core library, so a fact verified in the browser is reported as verified the next time an agent asks.}
\label{fig:arch}
\end{figure}
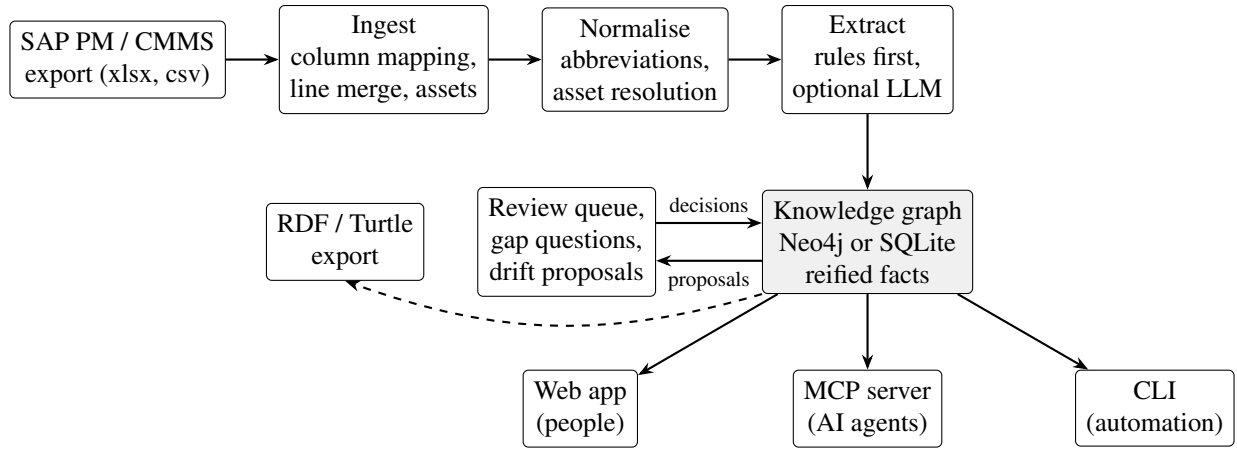

Figure~\ref{fig:arch} shows the data flow. Uploaded files are immutable evidence: people never edit raw rows, only the knowledge derived from them. The graph is rebuilt from all registered source files; when an order appears in several files, the newest file wins. Rebuilds preserve verification decisions for any fact whose extracted value is unchanged, and technician-stated facts are carried over untouched.

\subsection{Ontology}

{\sloppy Table~\ref{tab:classes} lists the core classes. Relations include \code{locatedIn} (asset to process area), \code{onAsset}, \code{involvesComponent}, \code{observedSymptom}, \code{hasFailureMode}, \code{hasRootCause} and \code{resolvedBy} (event to term), \code{consumed} (event to spare part), \code{realizes} (spare part to component type), \code{typicallyAffects} (failure mode to component) and \code{inCategory} (root cause to cause category).\par}

\begin{table}[t]
\centering\small
\begin{tabularx}{\linewidth}{@{}l L@{}}
\toprule
Class & Role and examples \\
\midrule
ProcessArea & Top of the asset hierarchy: parboiling, paddy-to-rice milling, utilities, packing, warehouse \\
Asset & Maintainable item with a functional location, class and shop-floor aliases (e.g.\ \emph{Dryer elevator}, alias ``LSU elev'') \\
Component & Part of an asset: bearing, V-belt, sprocket, plummer block, rubber roll, mechanical seal \\
SparePart & SAP material that realises a component type, with material code and unit cost \\
Symptom & Observation: abnormal noise, vibration, belt slip, tripping, reduced output \\
FailureMode & ISO~14224-aligned, with mechanism code: bearing wear (2.4 Wear), belt slippage (1.5 Looseness), seal leakage (1.1 Leakage) \\
RootCause & Misalignment, inadequate lubrication, dust ingress, overload, foreign material, cavitation; grouped into cause categories \\
Action & Replace, repair, weld repair, realign, lubricate, adjust, refurbish, modify \\
MaintenanceEvent & One work order: dates, work type, cost, downtime, breakdown flag \\
Fact & Reified statement (subject, predicate, object) with confidence, method, status, verifier and evidence \\
Evidence & A source row and field, or a technician statement \\
\bottomrule
\end{tabularx}
\caption{Core ontology classes. The vocabulary contains 26 components, 20 failure modes, 13 root causes, 10 actions and 9 symptoms, each with synonyms used for matching.}
\label{tab:classes}
\end{table}

Reification is the central design choice. A statement such as ``order 30000230188 has failure mode \emph{bearing wear}'' is stored as a fact with: a confidence in $[0,1]$; the extraction method (\code{structured}, \code{rules}, \code{inferred}, \code{llm}, \code{llm+rules}, \code{technician}, with an \code{+edited} suffix after human correction); the source field and the exact words it was read from; and a status that moves from \emph{proposed} to \emph{verified} or \emph{rejected}, with the verifier's name and time. In the RDF export, verified facts are additionally asserted as direct triples, so plain SPARQL queries see only trusted knowledge while proposed facts remain available in reified form. In Neo4j, facts are relationship properties, and every node carries a \code{:PM} label so the system never touches other data in a shared database.

\subsection{Technical language processing pipeline}

\paragraph{Ingestion.} Column names are matched to canonical fields by exact lookup, then fuzzy matching, and users can correct the mapping on screen; only order number, date and description are required. Date order (day-first or month-first) is inferred from the values. Rows are grouped by order: the modal description and work type are taken, header fields use the first non-empty value, and the earliest and latest posting dates are both kept. Rows without an order become consumption-only records. A data-quality flag is raised when a breakdown (from the breakdown column or from words such as ``broken'' in the text) carries the work type \emph{Preventive}.

\paragraph{Normalisation and asset resolution.} Text is lower-cased, letter--digit runs are split (``elev1'' $\rightarrow$ ``elev 1''), and a plant abbreviation table is applied (``parb'' $\rightarrow$ parboiling, ``brg'' $\rightarrow$ bearing, ``plmbr blck'' $\rightarrow$ plummer block, ``b/d'' $\rightarrow$ breakdown). Assets are resolved in order of reliability: a functional location that matches the register (confidence 0.98); otherwise the longest alias match in the normalised text, disambiguated by process-area cues (0.85 with an area cue, 0.80 without, 0.50 if still ambiguous); otherwise a fuzzy alias match (0.45). An asset-discovery step proposes new register entries from unresolved functional locations and from residual noun phrases after component, action and symptom words are stripped, accounting for SAP's 40-character truncation.

\paragraph{Rule-based extraction.} Sources are consulted in a fixed precedence: structured SAP columns, then technician remarks, then the short text, then the materials issued. Specific rules encode domain knowledge:
\begin{itemize}
  \item Values from structured columns that map to the vocabulary receive confidence 0.95; values that do not map become drift candidates.
  \item The asset's own name and aliases are removed before component matching, so ``Hodi belt conv'' is not read as a conveyor-belt component.
  \item A component named in the text and confirmed by an issued material receives 0.92; named but unconfirmed, 0.75; read only from materials, 0.75, preferring the highest-cost line and ignoring consumables such as adapter sleeves when a primary part is present.
  \item Root causes are read only from the root-cause column or technician remarks, never from the short text, where words such as ``stone'' usually name a component.
  \item A failure-mode phrase is accepted only if it is compatible with the extracted component (for example, \emph{belt slippage} requires a belt). An incompatible phrase is evidence of a failure the vocabulary cannot express on that component and becomes a drift candidate. With no phrase, the component's default mode is inferred at confidence 0.55 and labelled as inferred.
  \item Lubrication rounds, inspections and modifications without a symptom are classified as planned work and receive no failure mode or root cause.
\end{itemize}

\paragraph{LLM as second reader.} When enabled, events are sent in batches to any OpenAI-compatible endpoint (a hosted provider or a local model). The prompt constrains output to vocabulary identifiers, or \code{other:<phrase>} when the text clearly states something outside the vocabulary, and requires a per-field confidence and the exact evidence words. Responses are cached by a hash of the event payload, prompt version and model, so rebuilds do not repeat calls; malformed batches are split recursively, and any failure falls back to the rules for the affected events. The merge policy (Section~\ref{sec:llm}) was set from measurement:
\begin{enumerate}
  \item Structured columns always win; agreement raises confidence to 0.97.
  \item On disagreement the rule value is kept, its confidence drops by 0.15 (floor 0.3), and a note records the LLM's suggestion so the fact rises in the review queue.
  \item On agreement, confidence becomes $\min(0.93,\ \max(c_{\text{llm}}, c_{\text{rule}}) + 0.07)$.
  \item The LLM fills empty fields for failure mode, root cause, action and symptom at confidence $\min(0.7, c_{\text{llm}})$, but never components.
  \item An \code{other:} answer counts as drift only where the rules found nothing.
  \item The LLM cannot reclassify a failure as planned work when the rules found a component or symptom.
\end{enumerate}

\paragraph{Drift detection.} Phrases that fail to map, from structured columns, incompatible failure-mode evidence or LLM \code{other:} answers, are grouped into drift proposals with the orders they occur in. An engineer can approve a proposal as a new term (with label, synonyms, applicable components or cause category), map it as a synonym of an existing term, or reject it. Approved changes are written to an extension file and applied on the next build; the built-in vocabulary is never modified.

\subsection{Human verification and knowledge capture}

The review queue presents proposed facts lowest-confidence first, each with its evidence snippet, source field and extraction method, and allows a reviewer to verify, correct (which verifies the corrected value) or reject. Bulk verification above a confidence threshold is available and logged. Every decision is written to an audit log with the reviewer's name.

Knowledge-gap questions are generated where an asset has at least two events with the same failure mode and fewer than half of them have a confident or verified root cause. Each question names the asset, the failure mode, the number of unexplained events and the date range, asks what usually causes the failure and what check would confirm it, and suggests the technician who most often worked those orders. An answer that maps to a known cause becomes a proposed asset-level technician fact; an answer that names an unknown cause becomes a drift proposal.

\subsection{Reasoning}

All reasoning functions return the orders behind each result, a confidence, and a sparse-evidence flag when fewer than $\theta = 3$ events support the answer.

\paragraph{Root-cause ranking.} For an asset and an optional symptom or failure-mode filter, each matching failure event $e$ with a root-cause fact $f_e$ contributes weight $\omega(f_e)$ to that cause, where $\omega = 1$ for verified facts, the extraction confidence for proposed facts, and $0$ for rejected facts. Technician statements for the asset add $2.0$ if verified and $1.2$ if proposed. If the asset has fewer than $\theta$ explained events, events from assets of the same class are added at half weight. Scores are normalised, $s_c = w_c / \sum_{c'} w_{c'}$, and banded as high ($s_c \ge 0.5$), medium ($\ge 0.25$) or low, or as low (sparse evidence) when support is below $\theta$. The response also lists typical actions and parts and the number of unexplained events, pointing to the open knowledge-gap questions.

\paragraph{Similar failures.} Each failure event is indexed as a bag of tokens from its description, symptom, remarks, recorded root cause and extracted labels, after normalisation, and ranked against a free-text query with Okapi BM25 \citep{robertson2009bm25} ($k_1 = 1.2$, $b = 0.75$).

\paragraph{Bad actors, recurrence and spares.} Bad actors are a Pareto ranking of failure events by cost, downtime or frequency, grouped by asset, failure mode or both, reporting how many groups account for 80\% of the total. Recurrence uses a sliding window to find the largest set of events with the same failure mode on the same asset within $W$ days (default 60, at least three events). Minimum spare stock per material is $S = \lceil \mu L + z\,\sigma\sqrt{L} \rceil$, with $\mu$ and $\sigma$ the mean and standard deviation of monthly consumption, $L$ the lead time in months and $z = 1.65$; parts are ranked by the breakdown downtime of the orders that consumed them.

\subsection{Interfaces}

\begin{figure}[t]
\centering
\includegraphics[width=\linewidth]{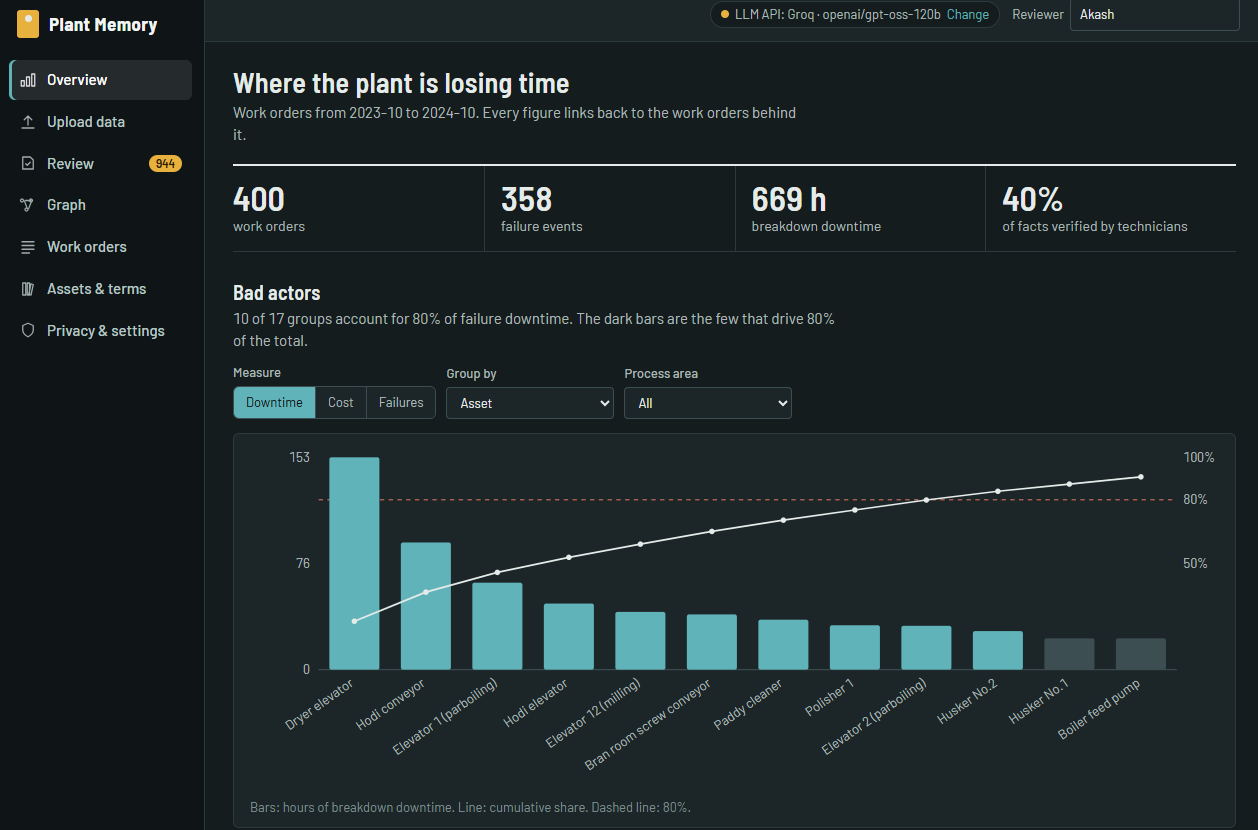}
\caption{Overview screen on the demo dataset: summary figures and a Pareto of breakdown downtime by asset, with the cumulative-share line and the 80\% threshold. The planted bad actor (the dryer elevator) ranks first.}
\label{fig:overview}
\end{figure}

\begin{figure}[t]
\centering
\includegraphics[width=\linewidth]{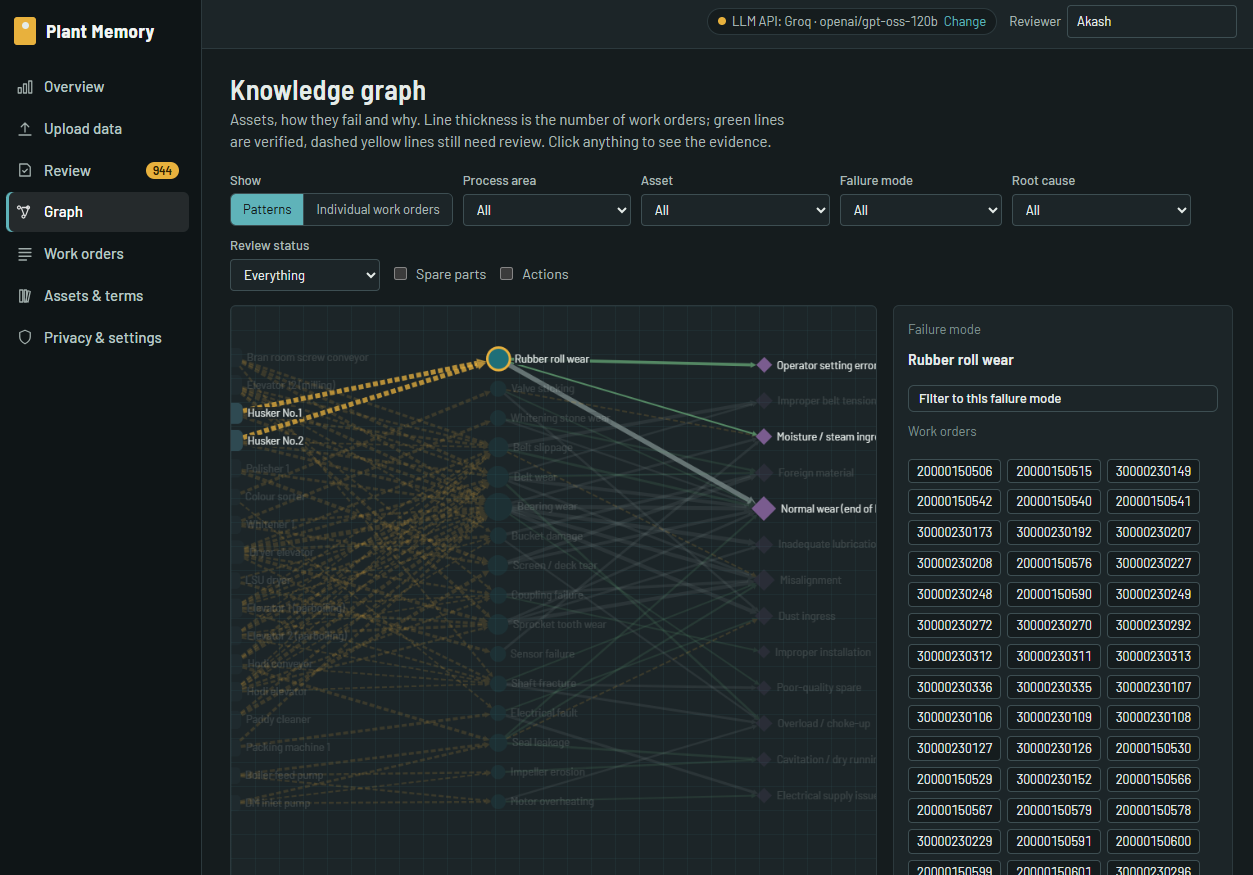}
\caption{Graph screen (patterns view): assets, failure modes and root causes, with edge width proportional to the number of work orders. Solid green edges are verified and dashed yellow edges await review. Selecting a node (here, \emph{rubber roll wear}) highlights its neighbourhood and lists the work orders behind it.}
\label{fig:graph}
\end{figure}

\begin{figure}[t]
\centering
\includegraphics[width=0.85\linewidth]{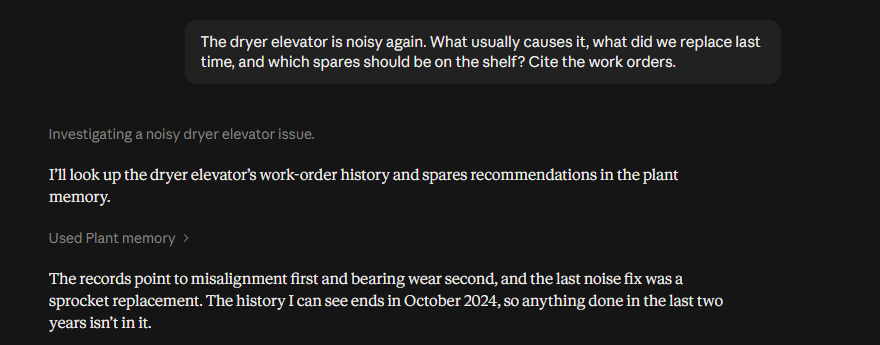}
\caption{A general-purpose assistant answering a planner's question through the Plant Memory MCP server. The answer is assembled from tool results over the verified graph and notes the limits of the available history.}
\label{fig:mcp}
\end{figure}

The MCP server exposes thirteen tools over standard input/output (Table~\ref{tab:mcp}) and the OWL schema as a resource. Its instructions tell the client to cite evidence orders, to describe proposed facts as unverified, and to recommend asking a technician when evidence is sparse. Asset arguments accept an identifier, a name or a shop-floor alias. The web application provides an overview (Figure~\ref{fig:overview}), an upload wizard with column mapping and asset discovery, the review queue, an interactive graph (Figure~\ref{fig:graph}), a searchable work-order table showing raw SAP lines beside extracted facts, an editor for the asset register and vocabulary, and privacy settings that state, before upload, whether work-order text will leave the machine. Figure~\ref{fig:mcp} shows a general-purpose assistant answering a planner's question through the MCP server.

\begin{table}[t]
\centering\small
\begin{tabularx}{\linewidth}{@{}l L@{}}
\toprule
Tool & Returns \\
\midrule
\code{list\_assets} & Assets by process area, with order and failure counts \\
\code{get\_asset\_history} & Orders for an asset with extracted facts, statuses, cost and downtime \\
\code{suggest\_root\_causes} & Ranked causes with scores, confidence bands, evidence orders and technician statements \\
\code{find\_similar\_failures} & BM25 matches for a free-text description, with causes, actions and parts \\
\code{rank\_bad\_actors} & Pareto by cost, downtime or frequency \\
\code{detect\_recurrence} & Repeat failures within a window \\
\code{recommend\_spares} & Minimum stock and criticality per material \\
\code{get\_evidence} & Raw SAP lines and facts for an order or fact \\
\code{data\_quality\_report} & Mislabelled work types, unresolved assets, split postings \\
\code{list\_knowledge\_gaps} & Open technician questions \\
\code{submit\_technician\_answer} & Records an answer as a \emph{proposed} fact (the only write tool) \\
\code{list\_drift\_proposals} & Terms awaiting an engineer's decision \\
\code{ontology\_terms} & The controlled vocabulary \\
\bottomrule
\end{tabularx}
\caption{MCP tools. Read tools are open to any connected client; the single write tool cannot change verified knowledge.}
\label{tab:mcp}
\end{table}

\section{Implementation}

Table~\ref{tab:stack} summarises the technology stack. The implementation is about 5,000 lines of Python; the web API, MCP server and CLI are thin layers over one core library. Storage is abstracted behind a repository interface with two implementations returning identical records: Neo4j, used when credentials are configured, and SQLite as an offline fallback and test reference. Graph queries use plain openCypher without plug-ins, so the same code is also exercised in tests on an embedded openCypher engine. Long builds run as background jobs with progress reporting. The system is packaged for \code{pip} and for Docker Compose with Neo4j.

\begin{table}[t]
\centering\small
\begin{tabularx}{\linewidth}{@{}l L@{}}
\toprule
Layer & Technology \\
\midrule
Core and pipeline & Python 3.10+, pandas, openpyxl, rule-based TLP, BM25 \\
LLM extraction & Any OpenAI-compatible API via httpx; evaluated with \code{openai/gpt-oss-120b} on Groq; local models via Ollama \\
Knowledge graph & Neo4j 5 (property graph, Cypher) or SQLite; RDF/Turtle export with rdflib; OWL~2 schema \\
Agent interface & Model Context Protocol server (Python MCP SDK, stdio transport) \\
Web API & FastAPI, Uvicorn, background build jobs \\
Web application & React 18, TypeScript, Vite, Cytoscape.js \\
Testing & pytest over SQLite, embedded openCypher and optional live Neo4j; API tests in an isolated workspace \\
Deployment & pip, Docker, Docker Compose \\
\bottomrule
\end{tabularx}
\caption{Technology stack.}
\label{tab:stack}
\end{table}

\paragraph{Privacy.} Maintenance records contain supplier names, costs and staff identifiers. The plant chooses one of three modes: local only (rules, no data leaves the machine), LLM API (text is sent to a hosted provider), or local LLM (an OpenAI-compatible model on the plant's own network). The active mode is displayed in the interface at all times.

\section{Evaluation}

\subsection{Data}

Real exports cannot be published, so evaluation uses a synthetic generator that reproduces the structure and messiness of the motivating export. It simulates 18 assets across five process areas for 12 months (October 2023 to October 2024) with Poisson failure arrivals per asset class and weighted scenarios per class (component, failure mode, symptoms, cause distribution, actions, quantities, downtime). Text is generated from deliberately messy templates, including abbreviations, upper-case variants, 40-character truncation and Hinglish remarks. Structured fields are sparsely populated: the root-cause column is filled for about half of breakdowns and a fifth of other failures, and work types are mislabelled on purpose. The generator plants four patterns: the dryer elevator fails at 2.2 times the class rate with misalignment over-represented; one V-belt is consumed across several elevators; husker rubber rolls dominate milling spend; and from the tenth month the Hodi conveyor shows gearbox oil leaks caused by a choked breather, a failure mode and cause absent from the vocabulary (drift). With seed 7 the export has 590 lines and 400 work orders (357 failures, 34 lubrication rounds and 9 modifications), plus 37 reservation lines. Ground truth is written to a separate file used only for evaluation.

\subsection{Extraction (rules only)}

Table~\ref{tab:e1} reports extraction against ground truth. Coverage is the share of events for which a value was extracted, and precision is accuracy on those. Root-cause coverage is limited by design: many work orders carry no cause evidence at all, and the system does not guess. Classification of failures versus planned work reaches precision 0.989 and recall 0.972. All 7 events whose true failure mode lies outside the vocabulary were flagged as drift.

\begin{table}[t]
\centering\small
\begin{tabular}{@{}lrrrrr@{}}
\toprule
Field & Events & Coverage & Precision & Accuracy & Macro F1 \\
\midrule
Asset & 400 & 1.000 & 1.000 & 1.000 & 1.000 \\
Component & 391 & 0.987 & 0.974 & 0.962 & 0.953 \\
Failure mode & 350 & 0.991 & 0.896 & 0.889 & 0.880 \\
Root cause & 350 & 0.620 & 1.000 & 0.620 & 0.805 \\
Action & 400 & 0.948 & 0.982 & 0.930 & 0.899 \\
Symptom & 357 & 0.608 & 0.982 & 0.597 & 0.767 \\
\bottomrule
\end{tabular}
\caption{Rule-based extraction on the synthetic export (400 work orders, seed 7).}
\label{tab:e1}
\end{table}

\subsection{Diagnosis}

We pose one question per (asset, failure mode) pair that has at least four true events and a unique dominant true cause: ``asset X keeps showing failure mode Y; what is the most likely root cause?'' This yields 28 questions. The baseline is what a planner obtains from the raw export: the most frequent value typed in the SAP root-cause column for those orders. Plant Memory answers with \code{suggest\_root\_causes} after a simulated review that verifies only facts read from structured columns (confidence $\ge 0.95$). Table~\ref{tab:e2} shows the results. In three of the five Plant Memory errors the system predicts \emph{normal wear} where the true dominant cause is inadequate lubrication or moisture ingress. In two of these, the most common value typed in the root-cause column was ``life over'' as well, so the recorded causes themselves pointed the wrong way; these are the situations the knowledge-gap questions are meant to resolve.

\begin{table}[t]
\centering\small
\begin{tabular}{@{}lccc@{}}
\toprule
Answerer & Top-1 & Top-3 & Answers citing evidence \\
\midrule
Raw SAP root-cause column & 0.536 & n/a & n/a \\
Plant Memory & \textbf{0.821} & \textbf{0.964} & \textbf{1.000} \\
\bottomrule
\end{tabular}
\caption{Diagnostic accuracy on 28 questions.}
\label{tab:e2}
\end{table}

\subsection{Rules versus LLM}
\label{sec:llm}

We ran \code{openai/gpt-oss-120b} with low reasoning effort on 399 of the 400 orders and compared it with the rules (Table~\ref{tab:llm}). As an independent extractor the LLM was less accurate on every field, most markedly on failure mode (55.4\% versus 88.9\%). A recurring error was reading the \emph{Preventive} work type literally and classifying genuine wear failures as planned work; in an initial merge that allowed the LLM to override the rules, this removed 99 failure events and reduced diagnosis top-1 accuracy from 82.1\% to 75.0\%. The LLM nevertheless added value where the rules were silent, correctly filling 8 actions and 5 symptoms (with 2 incorrect fills each), which motivated the second-reader policy described above. Under that policy, action accuracy rose from 93.0\% to 95.0\% and symptom accuracy from 59.7\% to 61.1\%, with component, failure-mode and diagnosis accuracy unchanged.

\begin{table}[t]
\centering\small
\begin{tabular}{@{}lccc@{}}
\toprule
Field & Rules & LLM alone & Rules + LLM (adopted policy) \\
\midrule
Component & 96.2 & 87.2 & 96.2 \\
Failure mode & 88.9 & 55.4 & 88.9 \\
Root cause & 62.3 & 46.9 & n/r \\
Action & 93.0 & 92.5 & 95.0 \\
Symptom & 59.7 & 53.5 & 61.1 \\
\midrule
Diagnosis top-1 & 82.1 & n/a & 82.1 \\
\bottomrule
\end{tabular}
\caption{Accuracy (\%) of rules, the LLM as an independent extractor, and the merged output. n/r: not reported separately; the LLM filled one root cause correctly where the rules were silent.}
\label{tab:llm}
\end{table}

\section{Discussion and limitations}

The evaluation is favourable to the system for a structural reason: the ground truth comes from the same generator as the data, and the abbreviation table and synonyms were written alongside the generator's vocabulary. Real exports will contain unseen abbreviations, other languages and plant-specific idioms, and accuracy will be lower. This is the case the review queue, drift proposals and vocabulary editor are designed for, but their effectiveness has not yet been measured. A credible evaluation requires 50 to 100 hand-labelled orders from a real plant, which would also show whether the rules-versus-LLM ranking holds once the rules face unfamiliar text. The comparison with an LLM is likewise specific to one model, one prompt and synthetic text.

Other limitations follow from scope. The system has been built for a single plant with one taxonomy; whether the ontology transfers to other equipment and industries is untested. There is no user study of technician engagement with verification, which is the main operational risk. The application is single-user, without authentication, and table and graph views slow down beyond roughly 50,000 lines. Ontology changes are audited but not yet versioned or reversible, and work-order drafting and preventive-maintenance interval suggestions remain future work.

\section{Conclusion}

Plant Memory treats maintenance text as evidence and maintenance knowledge as a set of claims that people verify. Rule-first technical language processing, an ISO~14224-aligned ontology with reified provenance, and a verification loop with knowledge-gap questions and drift proposals together produce a graph that both planners and AI agents can query, with every answer traceable to the work orders behind it. On synthetic data the approach substantially outperforms the raw root-cause column for diagnosis, and a measured comparison suggests that, for this kind of text, an LLM is more useful as a second reader than as the primary extractor. The next step is evaluation on real plant data with real technicians.

\section*{Code availability}
The implementation, synthetic data generator, evaluation harness and reports are available at \url{https://github.com/AkashRajSingh/Plant-Memory}. The repository was made public in October 2026.

\section*{Use of AI tools}
AI assistants were used during software development and in drafting and editing this manuscript. The author reviewed all content and is responsible for it. The LLM evaluated in Section~\ref{sec:llm} is part of the system under study.


\begin{thebibliography}{99}\small

\bibitem[Anthropic(2024)]{anthropic2024mcp}
Anthropic.
\newblock Introducing the Model Context Protocol.
\newblock \url{https://www.anthropic.com/news/model-context-protocol}, November 2024.

\bibitem[Bikaun et~al.(2024)]{bikaun2024maintie}
T.~K. Bikaun, T.~French, M.~Stewart, W.~Liu, and M.~Hodkiewicz.
\newblock {MaintIE}: A fine-grained annotation schema and benchmark for information extraction from maintenance short texts.
\newblock In \emph{Proceedings of LREC-COLING 2024}, pages 10939--10951, Torino, Italy, 2024.

\bibitem[Brundage et~al.(2021)]{brundage2021tlp}
M.~P. Brundage, T.~Sexton, M.~Hodkiewicz, A.~Dima, and S.~Lukens.
\newblock Technical language processing: Unlocking maintenance knowledge.
\newblock \emph{Manufacturing Letters}, 27:42--46, 2021.
\newblock doi:10.1016/j.mfglet.2020.11.001.

\bibitem[Dima et~al.(2021)]{dima2021adapting}
A.~Dima, S.~Lukens, M.~Hodkiewicz, T.~Sexton, and M.~P. Brundage.
\newblock Adapting natural language processing for technical text.
\newblock \emph{Applied AI Letters}, 2(3):e33, 2021.

\bibitem[Edge et~al.(2024)]{edge2024graphrag}
D.~Edge, H.~Trinh, N.~Cheng, J.~Bradley, A.~Chao, A.~Mody, S.~Truitt, and J.~Larson.
\newblock From local to global: A graph {RAG} approach to query-focused summarization.
\newblock arXiv:2404.16130, 2024.

\bibitem[Hodkiewicz and Ho(2016)]{hodkiewicz2016cleaning}
M.~Hodkiewicz and M.~T.-W. Ho.
\newblock Cleaning historical maintenance work order data for reliability analysis.
\newblock \emph{Journal of Quality in Maintenance Engineering}, 22(2):146--163, 2016.

\bibitem[ISO(2016)]{iso14224}
International Organization for Standardization.
\newblock {ISO 14224:2016} Petroleum, petrochemical and natural gas industries -- Collection and exchange of reliability and maintenance data for equipment, 2016.

\bibitem[Karray et~al.(2019)]{karray2019romain}
M.~H. Karray, F.~Ameri, M.~Hodkiewicz, and T.~Louge.
\newblock {ROMAIN}: Towards a {BFO} compliant reference ontology for industrial maintenance.
\newblock \emph{Applied Ontology}, 14(2):155--177, 2019.

\bibitem[Lewis et~al.(2020)]{lewis2020rag}
P.~Lewis, E.~Perez, A.~Piktus, F.~Petroni, V.~Karpukhin, N.~Goyal, H.~K{\"u}ttler, M.~Lewis, W.~Yih, T.~Rockt{\"a}schel, S.~Riedel, and D.~Kiela.
\newblock Retrieval-augmented generation for knowledge-intensive {NLP} tasks.
\newblock In \emph{Advances in Neural Information Processing Systems 33}, 2020.

\bibitem[Pan et~al.(2024)]{pan2024unifying}
S.~Pan, L.~Luo, Y.~Wang, C.~Chen, J.~Wang, and X.~Wu.
\newblock Unifying large language models and knowledge graphs: A roadmap.
\newblock \emph{IEEE Transactions on Knowledge and Data Engineering}, 36(7):3580--3599, 2024.

\bibitem[Robertson and Zaragoza(2009)]{robertson2009bm25}
S.~Robertson and H.~Zaragoza.
\newblock The probabilistic relevance framework: {BM25} and beyond.
\newblock \emph{Foundations and Trends in Information Retrieval}, 3(4):333--389, 2009.

\bibitem[Sexton et~al.(2017)]{sexton2017hybrid}
T.~Sexton, M.~P. Brundage, M.~Hoffman, and K.~C. Morris.
\newblock Hybrid datafication of maintenance logs from {AI}-assisted human tags.
\newblock In \emph{2017 IEEE International Conference on Big Data}, pages 1769--1777, 2017.

\bibitem[W3C(2012)]{owl2}
W3C OWL Working Group.
\newblock {OWL 2} Web Ontology Language document overview (second edition).
\newblock W3C Recommendation, 2012.

\bibitem[Woods et~al.(2024)]{woods2024ontology}
C.~Woods, M.~Selway, T.~Bikaun, M.~Stumptner, and M.~Hodkiewicz.
\newblock An ontology for maintenance activities and its application to data quality.
\newblock \emph{Semantic Web}, 15(2), 2024.
\newblock doi:10.3233/SW-233299.

\end{thebibliography}
\end{document}